\documentclass{llncs}
\usepackage[utf8]{inputenc}
\usepackage{graphicx}
\usepackage{amsmath}

\title{Automatic Lymphocyte Detection in H\&E Images with Deep Neural Networks}
\author{Jianxu Chen\inst{1}\thanks{This work was done when Jianxu Chen was an intern at Ventana Medical Systems, Inc.} \and Chukka Srinivas\inst{2}}
\institute{University of Notre Dame, United States \and Ventana Medical Systems, Inc.}

\begin{document}
\maketitle

\begin{abstract}
Automatic detection of lymphocyte in H\&E images is a necessary first step in lots of tissue image analysis algorithms. An accurate and robust automated lymphocyte detection approach is of great importance in both computer science and clinical studies. Most of the existing approaches for lymphocyte detection are based on traditional image processing algorithms and/or classic machine learning methods. In the recent years, deep learning techniques have fundamentally transformed the way that a computer interprets images and have become a matchless solution in various pattern recognition problems. In this work, we design a new deep neural network model which extends the fully convolutional network by combining the ideas in several recent techniques, such as shortcut links. Also, we design a new training scheme taking the prior knowledge about lymphocytes into consideration. The training scheme not only efficiently exploits the limited amount of free-form annotations from pathologists, but also naturally supports efficient fine-tuning. As a consequence, our model has the potential of self-improvement by leveraging the errors collected during real applications. Our experiments show that our deep neural network model achieves good performance in the images of different staining conditions or different types of tissues. 
\end{abstract}

\section{Introduction}

Immunetheropy with tumor infiltrated lymphocytes is an promising approach, and being widely investigated, for the treatment of cancers \cite{immuntherapy}. Detecting lymphocyte in H\&E stained histological tissue images is a critical step in the clinical studies. The quantification of lymphocytes provides a feasible solution to quantify the immune response, so that researchers can analyze the treatment outcome of immunetheropy quantitatively.

With the fast development of digital pathology, lymphocytes can be detected and examined by pathologists on computer screens with different visualization and annotation tools. However, the possible amount of lymphocytes in a single whole-slice (WS) image may range from tens to thousands, even maybe hundreds of lymphocytes in a small field of view (FOV). A reliable fully-automated lymphocyte detection system can make the study reproducible and faster by orders of magnitude.

In this work, we demonstrate the effectiveness of deep learning approaches in automatic lymphocyte detection in H\&E images. In particular, we take the following practical considerations to make our deep learning scheme effective in lymphocyte detection.

\begin{itemize}
    \item \textbf{Staining and tissue variations}: A noteworthy feature of H\&E stained histpathological images is the possibly large variation of staining conditions and tissue types. Our experiments show that our approach is robust to considerable staining and tissue variability. 
    
    \item \textbf{Prior knowledge}: The physical appearance of lymphocytes is a disk-like shape with diameter ranging from 14 to 20 microns. Given the image solution, it is straightforward to calculate the estimated size of lymphocytes in pixels. We take such important prior knowledge into account in the training process (see Section \ref{sec:train} for details).
    
    \item \textbf{Free-form supervision}: Labelling lymphocytes in H\&E images requires special expertise, which makes the annotation hard to crowd-source, like \cite{point}. So, the availability of ground truth is so limited that annotations should be expected free from any restricted form, such as a point around the center of lymphocytes, a point within similar non-lymphocyte objects (e.g., tumor cells or stromal cells), or even scribbles at non-lymphocyte pixels provided as negative examples. 
    
    \item \textbf{Human computer interaction for fine-tuning}: In case detection errors are found and corrected by pathologists, the model should be able to adapt to such ``new knowledge". By carefully designing the training scheme and supporting free-form supervision, our model can be easily fine-tuned through such human computer interaction. As a consequence, it make the entire model able to improve by itself in the progress of application in practice. 
\end{itemize}

In the literature, the first, and only, deep learning model for lymphocyte detection was discussed in \cite{cwu}, which employs a generic model to classify a small image patch as a lymphocyte or not. Formulating the problem as a patch-based CNN classification can result in an extremely long inference time and potentially lower accuracy than fully convolutional networks (FCN) \cite{fcn}. Such inferior performance could be the consequence of poor generality in case of limited training data. 

FCN has been widely used in medical image segmentation, such as \cite{unet,cumed,aaai} and even in 3D problems \cite{nips}. But, the exact object boundaries are less important than the detection of each object. Also, pixel-wise ground truth labels are nearly impossible to collect in our problem. Therefore, classic FCN formulated for segmentation is not directly applicable for lymphocyte detection.

In terms of object detection, FCN has been generalized to semantic object detection in images \cite{rcnn,faster}. In these methods, two sibling networks are trained. One is for the regression of the object bounding box, while the other one is to classify the object types. Given the fact that lymphocytes have relatively uniform sizes, we can adopt the idea in \cite{rcnn} but omitting the bounding box regression.

In this work, we propose to train an FCN to predict the probability of each pixel of being within a lymphocyte, which can be viewed as a model solving detection and classification in one shot. Our approach combines the ideas in the original FCN for segmentation and some new techniques \cite{unet,dropout,enet,resnet,allconv}. Our approach achieves promising results in real experiments (see Section \ref{sec:exp}).

\section{Methodology}
\label{sec:method}

In this section, we will describe the details of our deep learning model and the training strategy. Our model is extended from the fully convolutional network (FCN) proposed in \cite{fcn} by combining the ideas in \cite{unet,enet,resnet,allconv,dropout} so as to build an effective model for our problem. Moreover, we carefully design a new training scheme, which takes human prior knowledge into account and consequently can utilize free-form annotation efficiently and is capable of improving itself during real application. Finally, we will discuss pre-processing steps to prepare the input and the post-processing steps to generate the position and calculate the confidence score of each detected lymphocyte.

\subsection{Network Architecture}
\label{sec:architecture}

\begin{figure}[htb]
    \centering
    \includegraphics[width=0.99\textwidth]{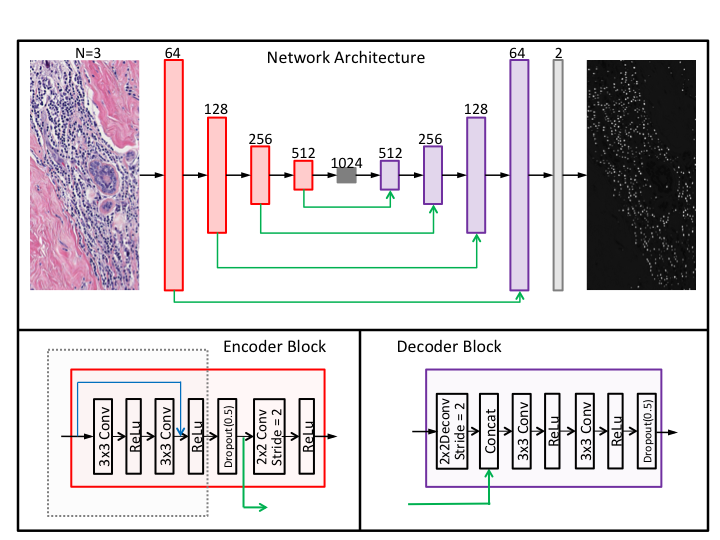}
    \caption{The overall architecture of the whole network and two key elements, encoder blocks and decoder blocks. The bridge block, i.e., the solid dark gray box in the middle of the network, has the same structure as part of the encoder block, labelled in the black dot box. The size of each block indicates the scale that the block works on. The number above each block is the number of feature maps in the block output. At the end, a $1\times 1$ convolution and softmax (see the light gray box) are performed to generate the probability map as the result. The blue arrow connector in the encoder block is the shortcut connection and the green arrow connectors are skip-layer connections (see Section~\ref{sec:architecture} for details).}
    \label{fig:net}
\end{figure}

The overall architecture of our proposed model is shown in Fig.~\ref{fig:net}. In essence, our model is an extention of the fully convolutional network (FCN) proposed in \cite{fcn}. For details of FCN, we refer \cite{fcn} for the full details and analysis. 

The whole network is formulated in an encoder-decoder framework. Each encoder block (rf. red boxes in Fig.~\ref{fig:net}), processes the image at a certain scale with a residual learning function (two $3\times 3$ convolutions and ReLU with a shortcut connection) and transforms the image to the upper scale (i.e., lower resolution) by a $2\times 2$ convolution with stride 2. Four consecutive encoder blocks can generate highly abstracted contexts, which are fed into the bridge block (rf. the solid dark gray box in Fig.~\ref{fig:net}). Then, the bridge block distills the highest level abstraction with a residual learning function (similar to a encoder block, but with no scale transformation function). With the extracted hierarchical information, the decoder blocks start to gradually restore the resolution one scale a time. At each scale, the decoder takes two inputs: the abstraction in the corresponding encoder block received through a skip connection (rf. green arrow connectors in Fig.~\ref{fig:net}) and the restored finer details from higher scale abstraction through a $2\times2$ deconvolution with stride 2. Then, the encoder block fuses information by two $3\times3$ convolutions and ReLU activations.  After four consecutive decoder blocks, the information is restored to the original resolution while the hierarchical features have been embedded in the feature maps. A $1\times1$ convolution and softmax function (rf. the light gray box in Fig.~\ref{fig:net}) are performed at the end to predict the probability of each pixel belonging to a lymphocyte.

Our network is an extension of FCN with the following four specific modifications:

\begin{itemize}
    \item Inspired by \cite{enet}, we formulate the FCN model as a encoder-decoder framework, which gradually decodes the information starting from the bridge block layer by layer.
    \item Inspired by \cite{unet}, we keep rich features (i.e., the same number of feature maps in the commensurate encoder blocks) in the decoder blocks, which we find very important in semantic segmentation in the medical context.
    \item Inspired by \cite{allconv}, the pooling layers in FCN are replaced by $2\times 2$ convolutions with stride 2 to perform down-sampling by half. 
    \item Inspired by \cite{resnet}, shortcut link is added in each encoder block to improve the effectiveness and efficiency of training, and therefore boost the performance in deep neural network.
    \item Inspired by \cite{dropout}, dropout layers are inserted in both the encoder and decoder blocks to avoid overfitting, which is very common in the medical domain. 
\end{itemize}

\subsection{Training}
\label{sec:train}

\textbf{Training data generation from annotation:} In the classic FCN model \cite{fcn} and most of its variations for semantic segmentation, fully labelled images are required for training, i.e. each pixel must be assigned a label. Recent work, such as \cite{point,imagelabel}, labelling one pixel for every objects (exhaustive) or assigning one label to each image can be used to train FCN in a weakly supervised fashion. But, such training data is extremely difficult to obtain for lymphocyte detection in H\&E images. To collect ground truth of lymphocyte positions, special expertise is necessary, due to other visually similar objects (e.g., certain tumor cells) and tissue and staining variability. Meanwhile, we expect the training data to contain a large variation, so we would like to include more sample images from different types of tissues or stained in different conditions. In fact, there could be tens to hundreds of lymphocytes in each FOV. To this end, labelling all pixels or even only one pixel for every lymphocytes exhaustively in a large number of images is labor-intensive and time-consuming, and may also easily introduce considerable noise.

We design a new training strategy which can effectively generate a large number of training data from a tiny amount of input from pathologists. The new training strategy also enables fine-tuning in the process of application by collecting error correction made by pathologists (see the next part). 

\begin{figure}[htb]
    \centering
    \includegraphics[width=0.7\textwidth]{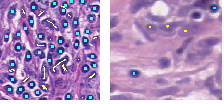}
    \caption{Examples of ground truth. Left: One example of the public data. The original labels of lymphocytes (dots in cyan) and the manually added negative samples (scribbles in yellow) are overlaid on the original H\&E image. Right: One example of the in-house data. The positive points (dots in cyan) and negative points (dots in yellow) are overlaid on the raw H\&E image. (All annotated dots are slightly dilated for clear visualization.)}
    \label{fig:cwu_train}
\end{figure}

To collect the ground truth, pathologist can make annotation on the images through a graphical user interface. There are two types of actions, most naturally actions, can be made: click and scribble. So, there are four types of annotations, as follows. Examples of the such annotations are shown in Fig.~\ref{fig:cwu_train}.
\begin{enumerate}
\item Positive point (PP): a single click around the center of a lymphocyte;
\item Positive scribble (PS): Strokes within a lymphocyte;
\item Negative point (NP): a single click within a non-lymphocyte object (visually similar to lymphocytes); 
\item Negative scribble (NS): Strokes either within a non-lymphocyte objects or in the background, especially in the areas between proximal lymphocytes.
\end{enumerate}

Next, we can build a label image, $L$, and a weight image, $W$, for each FOV. We perform a dilation for all pixels in $PP$ and $NP$. Let $PP_1$ and $PP_2$ (resp. $NP_1$) be the set of pixels dilated from $PP$ (resp. $NP$) by a disk of radius $r_1-5$ and $r_1$ (resp. $r_1+5$). Here, $r_1$ is a pre-determined parameter. All pixels in $PP_2\cup PS$ will have label 2 and all pixels in $NP_1\cup NN$ will have label 1. The remaining pixels will have label 0 (i.e., positions will not contribute to training).

For each pixel $p\in W$, $p$ is assigned the weight as:
\[ \begin{cases} 
      1, & p\in PP_1 \cup PS \cup NP \cup NP_1 \\
      0.5, & p\in PP_2 - PP_1 \\
      0, & Otherwise
   \end{cases}
\]

It is worth mentioning that prior knowledge can be incorporated to set the dilation parameter, i.e, a disk with radius $r_1$. Lymphocytes are round disk-like objects with diameter about 24 to 40 (in pixel). Therefore, we choose to use a disk template for dilation with $r_1$ set as 11 in our implementation. Such prior knowledge actually plays an important role in building the training data from ground truth, which enriches the limited pixel-level information in the annotation and implicitly imposes topological information. 

In each iteration, one FOV will be selected. The actual input to the deep learning model is a $K\times K$ patch created from the FOV, according to the following steps. First, we flip the image with $50\%$ probability ($25\%$ for horizontal flip and $25\%$ for vertical flip). Next, we rotate the image with $50\%$ probability, by a random angle $\theta$ ($\theta$ is a random integer from 1 to 360). Finally, we randomly select a position $(cx,cy)$ from all non-negative pixels in the label image and crop a patch centered at $(cx+\delta_x,cy+\delta_y)$ as the actual input. Here, $\delta_x$ and $\delta_y$ are random integers in $[-20, 20]$ meant to introduce randomness accounting for translation invariance. (Note: If the patch is partly out of the FOV, mirror padding is performed on the FOV. Also, the label image and the weight image will undertake the same transformation as the FOV.)

\textbf{Fine-tuning:} In the process of the application in practice, pathologists may find errors in the detection results. In this situation, one click on the screen through the user interface can actually provide a positive point or negative point. We can fine-tune the model periodically, say after every 200 points are collected. Fine-tuning can be conducted by using a small learning rate and a high momentum and following the aforementioned training procedure. Suppose there are $n$ FOVs, denoted as $F$ containing the newly collected ground truth. We randomly select two sets of $n$ FOVs, denoted as $A$ and $B$ from the previous training data. $F+A$ is used as the training data for fine-tuning and $B$ is used for validation. Here, the purpose of validation is to detect early stopping so that the model will not over-fit the new data.

\subsection{Pre-processing and post-processing}

We pre-process all the raw H\&E images using the stain normalization algorithm in \cite{stain}. Due to the nature of H\&E staining, it is important to normalize the data and also should be consistent for training and testing. 

After obtaining the probability map, we perform the following post-processing steps. 

(1) A binary mask is obtained from the probability map by a global threshold. In general, the threshold value is fixed for each trained model. In other words, we can select a proper value for the model after the training stage and fix it for application. When fine-tuning is performed later, the threshold value can be selected automatically so that the binary mask of an old training image is as close as possible to the binary mask of the same image before fine-tuning. 

(2) For each connected component in the binary mask, we compute the eccentricity, $e$, of each region, i.e., the eccentricity of the eclipse with the same second-moments as the region. If $e>0.8$ (an empirically determined parameter), the region will be discarded, considering the prior knowledge about the shape of lymphocytes. 

(3) Next, all regions whose size is not in $[S_1,S_2]$ will be removed. $S_1$ and $S_2$ are parameters indicate the estimated size of lymphocytes. 

(4) Finally, the centroid of each connected component with the binary mask will be returned as the position of lymphocytes. The confidence score of each detection is calculated as the average value of the corresponding region in the probability map.

\section{Experiments and Evaluations}

\subsection{Implementation Details}
\label{sec:implementation}

Our deep learning system is implemented in Matlab with MatConvNet \cite{matconvnet}. NVIDIA Quodro 4000 (2GB memory) is used for GPU acceleration. The network is initialized using the method in \cite{initialization} and trained from scratch. The training loss is cross-entropy and optimized with stochastic gradient descent, with batch size one and the weight of L2 regulation setting as $1e-5$. The learning rate and momentum used during the training are listed in Table \ref{tab:lr}.

\begin{table}[tb]
\centering
\begin{tabular}{| c || c | c |}
  \hline			
  Epoch & Learning Rate & Momentum \\
  \hline
  1-50 & 1e-4 & 0.9 \\
  51-120 & 1e-5 & 0.99 \\
  121-200 & 1e-6 & 0.999 \\
  \hline  
\end{tabular}
\caption{The learning rate and momentum used in the training process.}
\label{tab:lr}
\end{table}

\subsection{Ground Truth Collection}

The ground truth for problems in digital pathology can be limited and restricted due to the special expertise and intensive labor for annotation as well as the privacy issue. We collect ground truth for training from two sources. 

One is the lymphocyte detection dataset released by \cite{cwu}, which includes 100 FOVs. Each FOV is of $400\times400$ pixels (upsampled 4x from the original release). All lymphocytes are labelled exhaustively in each FOV (3064 lymphocytes in total). To create negative samples (i.e. non-lymphocyte positions), we manually scribble on the non-lymphocyte areas, especially the regions with similar appearances as lymphocytes (e.g. tumor cells) and the regions between proximal lymphocytes.

Besides, we include an in-house dataset collected from early breast cancer tissues, containing 99 FOVs, where each FOV is $8\sim15$ times larger than those in the public dataset. But, the lymphocytes are sparsely labelled (3770 lymphocytes in total). The negative samples are the positions of tumor cells and stromal cells annotated manually and verified by pathologists (7467 tumor cells and 781 stromal cells in total). Because the images are of large size and sparsely labelled, $400\times400$ (to be consistent with the image size in the public data) overlapping patches are generated. A patch with no annotations within the center $200\times200$ region is discarded. Finally, 7335 valid image patches are obtained.

Considering our data is from different sources, the ground truth is utilized as follows. We fix the epoch size as 175 for training and 25 for validation. Validation is performed after each epoch to check early stopping and over-fitting. The data from both the public and the in-house are randomly partitioned with ratio 0.9 into training/validation sets. The exact numbers of the data size are shown in Table~\ref{tab:data}. In each iteration, one image is randomly selected from the training (resp. validation) set of either the public or the in-house data (alternatively every iteration) for training (resp. validation). By doing these, we are meant to balance the impact from both data sources on the training procedure.

\begin{table}[tb]
\centering
\begin{tabular}{| c || c | c |}
  \hline			
  Dataset & Training Set & Validation Set \\
  \hline
  Public & 90 & 10 \\
  In-House & 6600 & 735 \\
  \hline  
\end{tabular}
\caption{The size of training/validation set of the public and in-house data. It is worth mentioning that even though the number of images in the training set of the public data is much less than that of the in-house data, the actual training received from both data is comparable, considering that the public data is exhaustively labelled and data augmentation (see Section~\ref{sec:train}) is used in each iteration.}
\label{tab:data}
\end{table}

\subsection{Qualitative Results}
\label{sec:exp}

\begin{figure}[htb]
    \centering
    \includegraphics[width=0.9\textwidth]{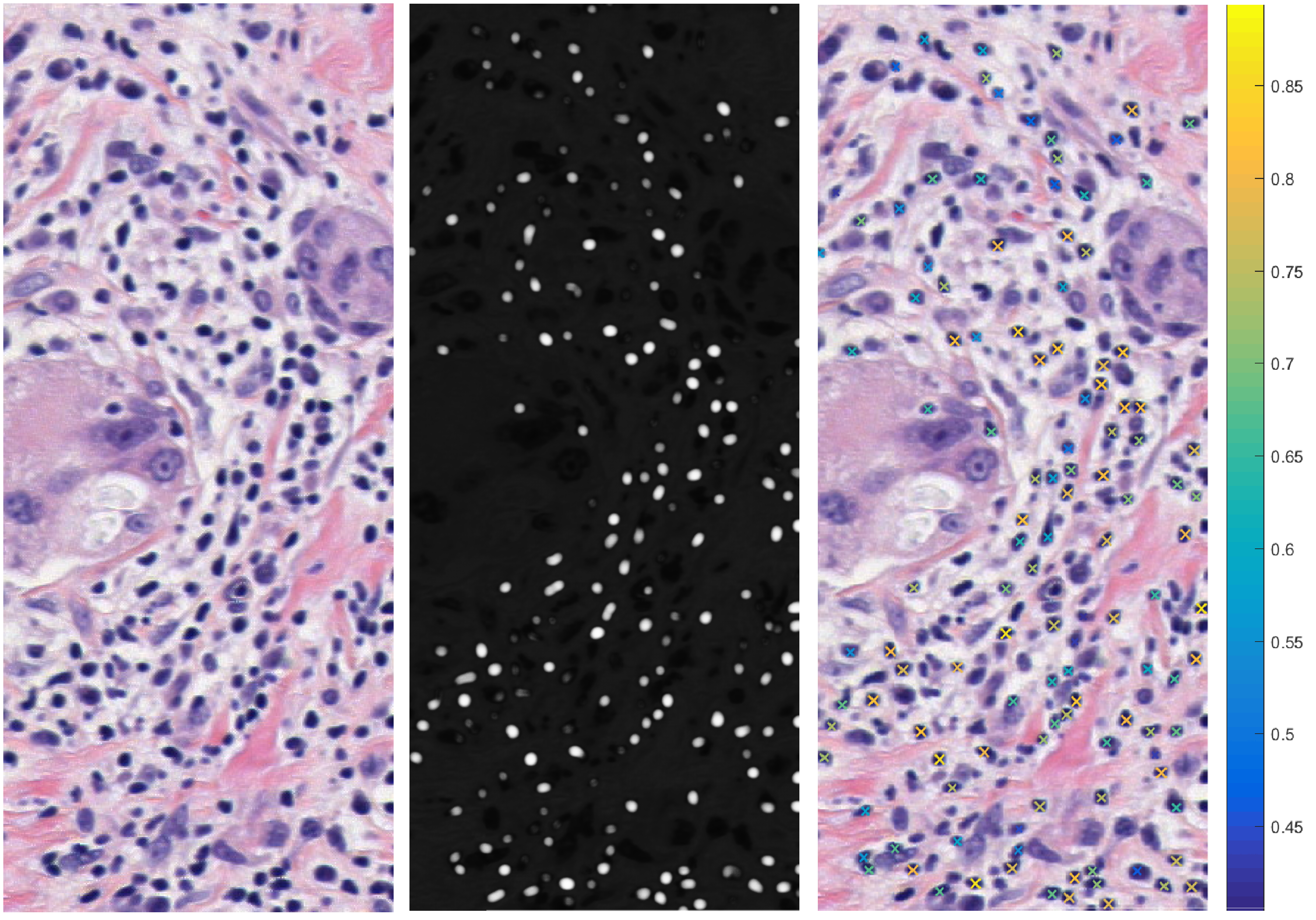}
    \caption{One example of detection results. Left: The raw H\&E image. Middle: The probability map generated by the deep learning model. Right: The visualization of the locations and confidence scores after post-processing. The color bar shows the confidence score (a real value from 0 to 1) of each detection.}
    \label{fig:result0}
\end{figure}

Due to the lack of large amount of ground truth annotation for evaluation, we only perform qualitative assessment in the current work and leave the extensive quantitative evaluation to the future work as more ground truth is being collected and supposes to take much more time. One example of detection results is demonstrated in Fig.~\ref{fig:result0}. The model generates a probability map and the post-processing step (as discussed in Section \ref{sec:method}) is applied to produce the location and confidence score of the detected lymphocyte. 

\begin{figure}[htb]
    \centering
    \includegraphics[width=0.9\textwidth]{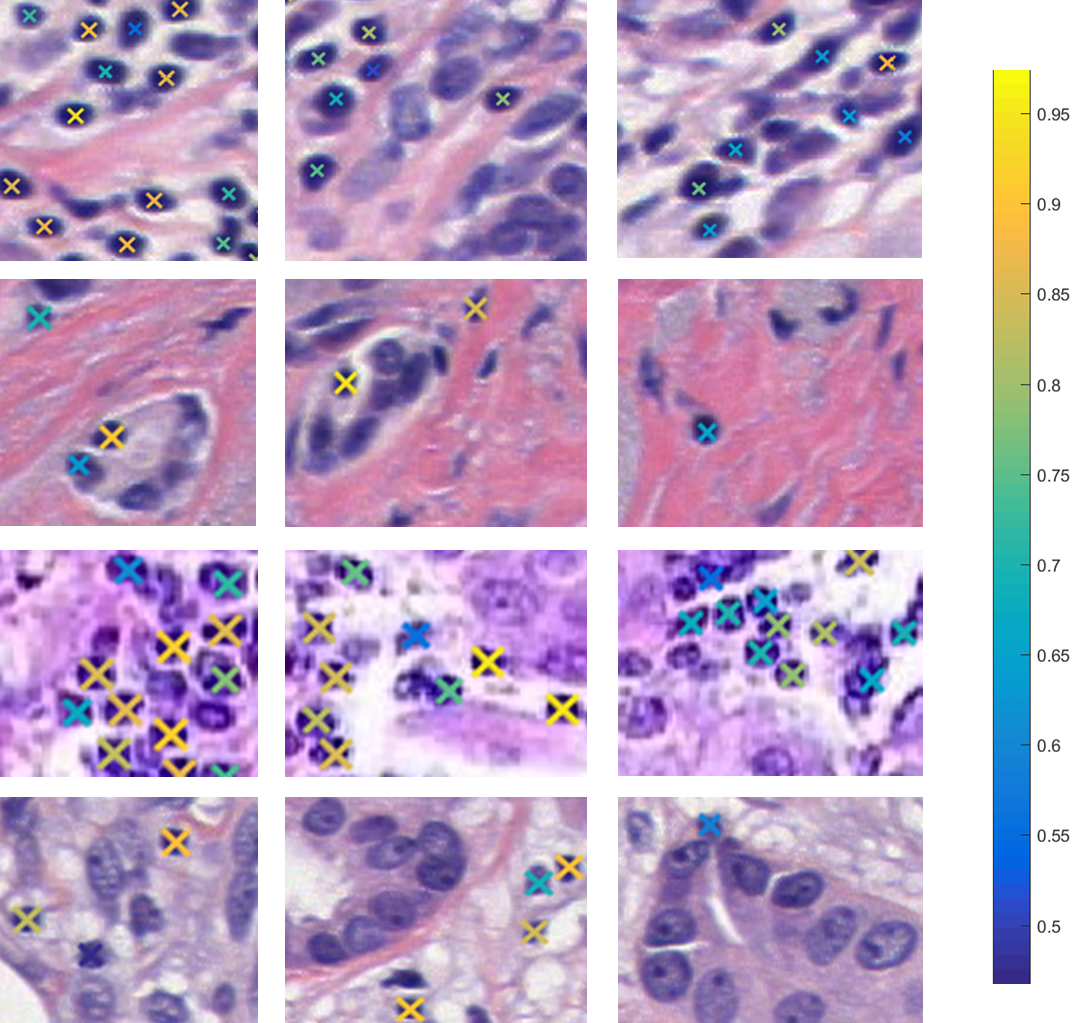}
    \caption{Detection results on different H\&E images. Row 1 is an image where considerable amount of lymphocyte exist. Row 2 is an image of a sample with lots of connected tissues. Row 3 is an image with relatively poor staining quality. Row 4 is an image containing mostly tumor cells. The color has the same indication of confidence scores as in Fig.~\ref{fig:result0}.}
    \label{fig:result1}
\end{figure}

\textbf{Robustness to Stain or Tissue Variability}: Fig.~\ref{fig:result1} shows detection results in different images. To some extent, we can observe that the performance is not very sensitive to the different staining conditions or different types of tissues.   

\begin{figure}[htb]
    \centering
    \includegraphics[width=0.85\textwidth]{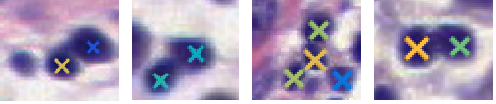}
    \caption{Detection results in the situation of proximal lymphocytes. The color has the same indication of confidence scores as in Fig.~\ref{fig:result0}.}
    \label{fig:result2}
\end{figure}

\textbf{Performance on Proximal Lymphocytes}: The lymphocytes may sometimes appear in clusters. Fig.~\ref{fig:result2} presents some sample results when two or more lymphocytes are close to each other or even with obscure separation boundaries. It is evident that our model is able to perceive the overall morphology and neighboring contexts to make predictions.

\section{Conclusions}

In this work, we develop a deep learning model for automatic lymphocyte detection. The model employs a new architecture extended from FCN by combining recent advances. The model is trained with a new strategy that efficiently utilizes free-form annotation. The new training scheme not only exploits the limited pathologists annotation efficiently, but also naturally enables the model self-taught by fine-tuning on the errors collected in the process of application. Experiements have shown that our model achieves promising results in H\&E images with large tissue or staining variations.    

\bibliographystyle{splncs03}
\bibliography{main}

\begin{thebibliography}{10}
\providecommand{\url}[1]{\texttt{#1}}
\providecommand{\urlprefix}{URL }

\bibitem{point}
Bearman, A., Russakovsky, O., Ferrari, V., Li, F.: What's the point: Semantic
  segmentation with point supervision. arXiv preprint arXiv:1506.02106  (2015)

\bibitem{aaai}
Chen, H., Qi, X., Cheng, J., Heng, P.: Deep contextual networks for neuronal
  structure segmentation. In: Proc. of AAAI Conference on Artificial
  Intelligence (2016)

\bibitem{cumed}
Chen, H., Qi, X., Yu, L., Heng, P.: Dcan: Deep contour-aware networks for
  accurate gland segmentation. arXiv preprint arXiv:1604.02677  (2016)

\bibitem{nips}
Chen, J., Yang, L., Zhang, Y., Alber, M., Chen, D.: Combining fully
  convolutional and recurrent neural networks for 3d biomedical image
  segmentation. arXiv preprint arXiv:1609.01006  (2016)

\bibitem{rcnn}
Girshick, R.: Fast r-cnn. In: Proc. of IEEE Conf. on Computer Vision (ICCV).
  pp. 1440--1448 (2015)

\bibitem{initialization}
Glorot, X., Bengio, Y.: Understanding the difficulty of training deep
  feedforward neural networks. In: Aistats. vol.~9, pp. 249--256 (2010)

\bibitem{resnet}
He, K., Zhang, X., Ren, S., Sun, J.: Deep residual learning for image
  recognition. In: Proc. of IEEE Conf. on Computer Vision and Pattern
  Recognition (CVPR) (2016)

\bibitem{cwu}
Janowczyk, A., Madabhushi, A.: Deep learning for digital pathology image
  analysis: A comprehensive tutorial with selected use cases. Journal of
  Pathology Informatics  7(1), ~29 (2016)

\bibitem{allconv}
J.T., S., Dosovitskiy, A., Brox, T., Riedmiller, M.: Striving for simplicity:
  The all convolutional net. arXiv preprint arXiv:1412.6806  (2014)

\bibitem{fcn}
Long, J., Shelhamer, E., Darrell, T.: Fully convolutional networks for semantic
  segmentation. In: Proc. of IEEE Conf. on Computer Vision and Pattern
  Recognition (CVPR). pp. 3431--3440 (2015)

\bibitem{stain}
Macenko, M., Niethammer, M., Marron, J., Borland, D., Woosley, J., Guan, X.,
  Schmitt, C., Thomas, N.: A method for normalizing histology slides for
  quantitative analysis. In: IEEE International Symposium on Biomedical
  Imaging: From Nano to Macro. pp. 1107--1110 (2009)

\bibitem{enet}
Paszke, A., Chaurasia, A., Kim, S., Culurciello, E.: Enet: A deep neural
  network architecture for real-time semantic segmentation. arXiv preprint
  arXiv:1606.02147  (2016)

\bibitem{imagelabel}
Pathak, D., Krahenbuhl, P., Darrell, T.: Constrained convolutional neural
  networks for weakly supervised segmentation. In: Proc of IEEE Conf. on
  Computer Vision (ICCV). pp. 1796--1804 (2015)

\bibitem{faster}
Ren, S., He, K., Girshick, R., Sun, J.: Faster r-cnn: Towards real-time object
  detection with region proposal networks. In: Advances in neural information
  processing systems. pp. 91--99 (2015)

\bibitem{unet}
Ronneberger, O., Fischer, P., Brox, T.: U-net: Convolutional networks for
  biomedical image segmentation. In: Proc. of Medical Image Computing and
  Computer-Assisted Intervention (MICCAI), Part III. pp. 234--241 (2015)

\bibitem{immuntherapy}
Rosenberg, S., Spiess, P., Lafreniere, R.: A new approach to the adoptive
  immunotherapy of cancer with tumor-infiltrating lymphocytes. Science
  233(4770),  1318--1321 (1986)

\bibitem{dropout}
Srivastava, N., Hinton, G., Krizhevsky, A., Sutskever, I., Salakhutdinov, R.:
  Dropout: a simple way to prevent neural networks from overfitting. Journal of
  Machine Learning Research  15(1),  1929--1958 (2014)

\bibitem{matconvnet}
Vedaldi, A., Lenc, K.: Matconvnet -- convolutional neural networks for matlab.
  In: Proceeding of the {ACM} Int. Conf. on Multimedia (2015)

\end{thebibliography}

\end{document}